\documentclass{article}

\usepackage{arxiv}

\usepackage[utf8]{inputenc} 
\usepackage[T1]{fontenc}    
\usepackage{hyperref}       
\usepackage{url}            
\usepackage{booktabs}       
\usepackage{amsfonts}       
\usepackage{nicefrac}       
\usepackage{microtype}      
\usepackage{lipsum}
\usepackage{graphicx}
\graphicspath{ {./images/} }
\usepackage{listings}
\usepackage{color} 
\definecolor{codegreen}{rgb}{0,0.6,0}
\definecolor{codegray}{rgb}{0.5,0.5,0.5}
\definecolor{codepurple}{rgb}{0.58,0,0.82}
\definecolor{backcolour}{rgb}{0.95,0.95,0.92}

\lstdefinestyle{mystyle}{
	backgroundcolor=\color{backcolour},   
	commentstyle=\color{codegreen},
	keywordstyle=\color{magenta},
	numberstyle=\tiny\color{codegray},
	stringstyle=\color{codepurple},
	basicstyle=\ttfamily\footnotesize,
	breakatwhitespace=false,         
	breaklines=true,                 
	captionpos=b,                    
	keepspaces=true,                 
	numbers=left,                    
	numbersep=5pt,                  
	showspaces=false,                
	showstringspaces=false,
	showtabs=false,                  
	tabsize=2
}

\lstdefinelanguage{json}{
	basicstyle=\normalfont\ttfamily,
	numbers=left,
	numberstyle=\scriptsize,
	stepnumber=1,
	numbersep=8pt,
	showstringspaces=false,
	breaklines=true,
	frame=lines,
	backgroundcolor=\color{backcolour},
	stringstyle=\color{codepurple},
	literate=
	*{0}{{{\color{codepurple}0}}}{1}
	{1}{{{\color{codepurple}1}}}{1}
	{2}{{{\color{codepurple}2}}}{1}
	{3}{{{\color{codepurple}3}}}{1}
	{4}{{{\color{codepurple}4}}}{1}
	{5}{{{\color{codepurple}5}}}{1}
	{6}{{{\color{codepurple}6}}}{1}
	{7}{{{\color{codepurple}7}}}{1}
	{8}{{{\color{codepurple}8}}}{1}
	{9}{{{\color{codepurple}9}}}{1}
	{:}{{{\color{codepurple}:}}}{1}
	{,}{{{\color{codepurple},}}}{1}
	{\{}{{{\color{codepurple}\{}}}{1}
	{\}}{{{\color{codepurple}\}}}}{1}
	{[}{{{\color{codepurple}[}}}{1}
	{]}{{{\color{codepurple}]}}}{1},
}

\title{Conversational AI Multi-Agent Interoperability\\[1ex] \large Universal Open APIs for Agentic Natural Language Multimodal Communications}

\author{ \href{https://orcid.org/0009-0008-7513-1255}{\includegraphics[scale=0.06]{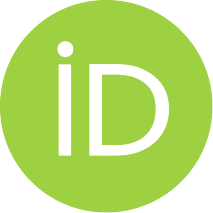}\hspace{1mm}Diego Gosmar}\thanks{Lead Author} \\
	Chief AI Officer XCALLY\\
	Open Voice Interoperability Initiative Member\\
	Linux Foundation AI \& Data\\
	Torino, TO 10100, Italy \\
	\texttt{diego.gosmar@ieee.org} \\
	\And
	\href{https://orcid.org/0000-0002-3389-2784}{\includegraphics[scale=0.06]{orcid.pdf}\hspace{1mm}Deborah A. Dahl} \\
	Principal Conversational Technologies\\
	Open Voice Interoperability Initiative Member\\
	Linux Foundation AI \& Data\\
	Plymouth Meeting, Pennsylvania, USA \\
	\texttt{dahl@conversational-technologies.com} \\
	\And
	\href{https://orcid.org/0009-0001-3770-4963}{\includegraphics[scale=0.06]{orcid.pdf}\hspace{1mm}Emmett Coin} \\
	Founder ejTalk\\
	Open Voice Interoperability Initiative Member\\
	Linux Foundation AI \& Data\\
	Bellville, Michigan, USA \\
	\texttt{emmett@ejtalk.com} \\
}

\begin{document}
\maketitle
\begin{abstract}
This paper analyses Conversational AI multi-agent interoperability frameworks and describes the novel architecture proposed by the Open Voice Interoperability initiative (Linux Foundation AI \& DATA), also known briefly as OVON (Open Voice Network). The new approach is illustrated, along with the main components, delineating the key benefits and use cases for deploying standard multi-modal AI agency (or agentic AI) communications.\\
Beginning with Universal APIs based on Natural Language, the framework establishes and enables interoperable interactions among diverse Conversational AI agents, including chatbots, voicebots, videobots, and human agents. Furthermore, a new Discovery specification framework is introduced, designed to efficiently look up agents providing specific services and to obtain accurate information about these services through a standard Manifest publication, accessible via an extended set of Natural Language-based APIs. The main purpose of this contribution is to significantly enhance the capabilities and scalability of AI interactions across various platforms.\\
The novel architecture for interoperable Conversational AI assistants is designed to generalize, being replicable and accessible via open repositories.
\end{abstract}

\keywords{Conversational AI \and Artificial intelligence \and AI Interoperability \and Multi-Agency \and Agentic AI \and Agentive AI \and Chatbot \and Voicebot \and Intelligent Assistant}

\section{INTRODUCTION}
As of this writing, the number of Chatbots and Voicebots available worldwide has reached figures above 200,000. For instance, a particular Chatbot Directory website reports\cite{chatbottle} a total of over 183,000 bots related to Facebook Messenger and Amazon Alexa Skills alone.\\
According to an analysis by Tidio.com\cite{tidio} roughly 1.5 billion people are using chatbots. The number of chatbot users worldwide is expected to continue growing, and by 2027, chatbots are projected to become the primary customer service channel for a quarter of businesses. This dramatic proliferation and commercial importance of Chatbots and Voicebots means that it is imperative to consider the overall worldwide ecosystem of Chatbots and Voicebots in order to improve how it can be made more efficient and effective.\\
For the remainder of this document, the term "agent" will be used to refer to an entity with the capacity to act, while "agency" or “agentic” will denote the exercise or manifestation of this capacity, in accordance with the definition provided by Markus Schlosser\cite{plato2015}.\\
Currently, Conversational AI encompasses interactions between human agents and AI agents, as well as an increasing number of interactions occurring directly among AI agents themselves. This research\cite{gosmarhyper} analyses how Conversational AI interactions are becoming pervasive and, in some use cases, nearly indistinguishable from human-to-AI interactions, particularly in terms of technical user experience and channels.\\
According to Cyberprotection Magazine\cite{cyber}, only 4\% of web content is accessible through search engines. This implies that 96\% of web content is not available to public generative AI large language models (LLMs). Consequently, AI agents will need to communicate with each other outside of public LLMs, especially when they need to escalate issues or request specific information available only from vertical and specialized AI assistants.\\
As a consequence, the proliferation of fragmented and specialized AI agents needing to communicate with each other, but without an interoperable framework, could lead to implementation complexity, duplication of effort, and ultimately user frustration. Defining interoperable standards for Conversational AI agent interactions will help mitigate these potential issues and increase scalability. For those reasons, the Open Voice Interoperability Initiative has defined a set of Universal APIs based on the Interoperable Conversation Envelope Specifications\cite{ovonconv}.\\
The rest of the paper is structured as follows.  Section 2 describes earlier related work and how it differs from the work described in this paper. Section 3 describes an open source approach to intelligent assistant interoperability. Section 4 provides some state diagrams that illustrate the life cycle of an exchange among interoperable agents. We review two detailed use cases in Section 5. Section 6 describes future directions and Section 7 covers our conclusions.

\section{PREVIOUS WORK}
In this section we review other investigations into ways to enable independent conversational assistants to exchange information. There are three previous threads of work addressing the problem of collaborating independent conversational assistants.\\
The first approach does not strictly involve independent assistants but independent modality components that collaborate, similarly to independent assistants, using a standard API. These modality components perform dialog functions such as speech recognition, natural language understanding, text-to-speech and so on. Examples of these kinds of architectures are the W3C Multimodal Architecture\cite{w3c} and the Galaxy Communicator Software Infrastructure\cite{galaxy}. We mention these in order to clarify how this previous work differs from the work we describe in this paper.\\
The second approach (Agentic AI) enables independent assistants to collaborate by hard-wiring the assistants together at development time. Examples of this are AutoGen\cite{autogen} and OpenDevin\cite{opendevin}. This approach is more flexible than developing a single system that can handle all of the anticipated functions, because new functionalities can be added by adding new assistants. This is especially useful for coding development and testing, i.e., having different LLMs performing specific tasks like software development and other LLMs “reflecting” on the result to perform test simulations, fix issues and improve the results. However, this approach has several drawbacks. It still requires all of the collaborating assistants to be known ahead of time, it requires the developer to be familiar with all of their APIs based on specific coding instead of natural language, and it requires all of the assistants to be based on a common technology; in this case LLMs. As a result, with this approach it would be quite complex to deploy interoperable Conversational AI assistants capable of handing over voice, video, and chat interactions from humans to different levels of independent AI agents specialised in Natural conversations.\\
Another system for supporting the interaction of independent assistants is VoiceXML\cite{vxml}. Very simple collaboration among voice dialog systems was made possible with the VoiceXML <transfer> element, which instructed the platform to transfer a user’s call to another location, either to a phone number or, more generally, to a URI. The transfer could be either a “bridge” transfer that would return the call to the originating VoiceXML platform at the end, or a “blind” transfer which did not allow the original platform to resume the call. VoiceXML differs  from the work described here in that a VoiceXML transfer requires that Agent B, the receiving agent, was either another VoiceXML application or a human agent, while this paper proposes that conversational assistants with any internal architecture should be able to collaborate as long as they adhere to the communication standards.\\
In the third approach, fewer assumptions are made about the internal structures of the assistants. For example, in the Open Agent Architecture\cite{oaa}, it is only necessary that the assistants adhere to the Inter Agent Communication Language (ICL); their internal architecture is not important. However, the messages sent between agents are in the form of a request for the secondary agent to do something, which requires the requesting agent to have good information about the secondary agent’s capabilities. Moreover, secondary agents need to have the ability to interpret ICL, which is a fairly fine-grained semantic representation.\\
In contrast, the work to be presented here, the fourth approach, for the most part simply passes on the user’s request to the secondary assistant, which minimally only needs to know how to interpret natural language, although agents can also communicate using structured formats that they both understand. In contrast to the ICL, the Open Voice Conversation Envelope API can also send information to a human agent at the endpoint.\\
The DARPA Communicator program\cite{darpa} also supported collaboration among multiple dialog managers, but the overall communication was coordinated through a central Hub, while in our approach all of the conversational assistants are independent.\\
In summary, while the goal of enabling multiple independent assistants to work together has been explored in previous work, no previous work has allowed independent assistants, potentially using totally different technologies (Generative AI LLMs, LMM, LAM, Non-Generative AI, or even no AI at all in some cases) and different channels (voice, video, text), to be as loosely coupled as the work presented here. We believe this an important feature because it greatly reduces the complexity of adding Open Voice integration to new assistants.\\
The Open Voice Specifications impose very few requirements on cooperating assistants (refer to Table~\ref{tab:table} for a detailed comparison of the different approaches).

\begin{table}[h!]
	\caption{Conversational AI Interoperability Collaboration Approaches}
	\centering
	\begin{tabular}{|p{4cm}|p{3cm}|p{3cm}|p{3cm}|}
		\hline
		\textbf{Approach} & \textbf{Description} & \textbf{Benefits} & \textbf{Drawbacks} \\
		\hline
		First Approach: Independent Modality Components & Independent modality components collaborate using a standard API (e.g., W3C Multimodal Architecture, Galaxy Communicator Software Infrastructure) & Modular design, Components perform specific dialog functions & Not fully independent assistants, Limited to modality components\\
		\hline
		Second Approach: Agentic Hard-Wired Collaboration & Independent assistants are hard-wired together at development time (e.g., AutoGen, OpenDevin). Uses LLMs for specific tasks and "reflection" for testing and improvement. & Flexible addition of new functionalities, Specialized LLMs for specific tasks & All assistants must be known ahead of time, Requires familiarity with all APIs, Based on common technology (LLMs) \\
		\hline
		Third Approach: Inter Agent Communication Language (ICL) & Assistants adhere to ICL, focusing on requests and task performance (e.g., Open Agent Architecture). & Less dependency on internal structures, Standard communication protocol & Requires detailed knowledge of secondary agent's capabilities, Need to interpret fine-grained ICL semantics\\
		\hline
		Fourth Approach: Open Voice Conversation Envelope API & Passes user requests to secondary assistants, which interpret natural language. Supports different technologies and channels (voice, video, text). & Very loosely coupled assistants, Supports various technologies (Generative AI, LMM, LAM, Non-Generative AI, etc.), Minimal requirements for integration, Simplified addition of new assistants & Relies on natural language interpretation, which may vary in accuracy\\
		\hline
	\end{tabular}
	\label{tab:table}
\end{table}

\section{CONVERSATIONAL AI INTEROPERABILITY}
The Open Voice Interoperability initiative has released the first version of the Interoperable Conversation Envelope Specification, with further updates available here\cite{ovonconv}.\\
This work is motivated by three simple premises:   First, that the ubiquity of foundation language models (FLMs) will lead to a multiplicity of specialised conversational agents instead of a monolithic all-knowing AI. Second, that these conversational agents will be much more effective if they can collaborate with each other to achieve tasks, being potentially hosted anywhere and developed with any kind of technology.  Third, that this collaboration will itself be based on unstructured natural language instead of on complex and specific coding.  In short, we anticipate that interaction between language-capable AIs will look remarkably like collaboration between people.\\
These premises are, of course, open to debate but, if they prove to be correct, there will be a need for interoperability between AIs that use different underlying architectures and technologies.  They may even speak different languages.  In the long-term it may even be possible for this interoperability to use the same mechanisms that people use to interoperate - namely email, instant messaging, social media, publications and even phone calls and video calls.    This will only become possible however, when conversational AI has reached human-level intelligence. In the near term, the Interoperable Conversation Envelope Specification\cite{ovonspec} has been developed to facilitate interoperability between language-capable AIs with varying levels of intelligence and varying degrees of speciality.\\
In the Open Voice Interoperability framework, conversational envelope messages are sent from one agent to another agent, carrying events. The messages support the following event types:
\begin{itemize}
\item \textbf{Utterance}:  each agent can receive utterances and generate utterances in a conversation.   These are expressed in a standard format called dialog events\cite{dialog}, supporting simple text and extensible to support audio and other media in the future.
\item \textbf{Whisper}: each agent can receive and generate Whispers.  These are out-of-band utterances that can be sent 'behind the scenes' to facilitate cooperation without being verbalised in the conversation itself.  Whispers also use Dialog Events\cite{dialog}.
\item \textbf{Invite}: agents can invite other agents to join a conversation.  They can also choose to accept such invites and leave conversations.
\item \textbf{Bye}: agents can terminate the conversation by sending the Bye dialog event message.
\item \textbf{Request/Publish Manifest}: each agent provides a simple structured manifest on request which describes its capabilities using natural language and simple structured keywords.\cite{manifest}
\item \textbf{Find/Propose Assistant}: agents can ask other agents to recommend third-party agents to help with a specific task.  Agents can also ask other agents if they want to service a specific task.
\end{itemize}
Details of the actual formats of these events can be found in the appendices and the specifications.
The conversational envelope framework draws no distinction between human agents and automated agents. This underlines the insight above that language-capable AIs will interact with one another in much the same way as humans cooperate with each other.  It further anticipates that such cooperation may actually involve a web of human and artificial agents.   In the classic instance of a single human 'user' interacting with a single conversational AI then the user is simply represented by a proxy agent adopting the role of a Demanding Agent as described below.  All agents can initiate requests or respond to requests.  Chains and trees of agent interactions are possible and encouraged. The next section adds detail about the interactions among agents through a set of state diagrams showing how the events defined in the OVON specifications cause state transitions among interoperable assistants. 

\section{STATE DIAGRAM REPRESENTATION}
Let's define a state diagram representing AI agents moving from one status to another in the environment as a visual representation that illustrates the different states an AI agent can occupy and the transitions between these states based on specific events or conditions. This type of diagram is useful for modelling the behaviour of AI agents, showing how they react to various stimuli or inputs and change their state accordingly. In particular, for conversational AI agents, a state diagram represents the different states or modes the agent can be in during a conversation with a user or another AI agent, and how it transitions between these states based on the inputs and interactions with the counterpart. This representation is useful for mapping the OVON Interoperability Specifications to the dynamic statuses and standard messages exchanged. Note that the state diagrams shown here only describe states that are relevant to inter-agent collaboration.  In the course of performing the functions they’re designed for, agents can be in a wide variety of internal states that are not of concern for interoperability among agents.\\
\\
The key components of such a state diagram are as follows:\\
\textbf{States}: the different conditions or modes the conversational AI agent can be in during a conversation. Each state might represent a specific context or task the agent is handling.\\
\textbf{Transitions}: the changes from one state to another, triggered by specific user inputs, agent messages, or internal conditions.\\
\textbf{Events}/Triggers: the inputs or conditions that cause the AI agent to move from one state to another. These are often user or other agent inputs or system conditions.\\
\textbf{Initial State}: the starting point of the conversation, where the agent begins its interaction.\\
\textbf{Final State}: the end of the conversation or a termination state, indicating the conversation is complete.\\
\textbf{Actions}: behaviours or responses the AI agent produces during transitions or while in a specific state.\\
\\
Let's define a \textbf{Serving Agent} as an AI entity designed to provide services, information, or assistance to users or other agents. It operates by responding to requests, performing tasks, and delivering outputs based on its capabilities and the inputs it receives. In the context of conversational AI, a serving agent handles inquiries (according to the Interoperable Conversation Envelope Specifications), executes commands, and engages in interactions to fulfill the needs of the user or the system it serves.\\
On the other hand, a \textbf{Demanding Agent} is an AI entity that initiates requests or tasks, seeking services, information, or assistance from other agents or systems. It operates by sending queries, commands, or prompts to other agents to achieve its objectives. In the context of conversational AI, a demanding agent engages in interactions (according to the Interoperable Conversation Envelope Specifications) where it plays the role of requester, needing responses or actions from the serving agent.\\
It should be noted that the same agent can serve either of these roles at different times (serving or demanding agent).

\subsection{Interoperable Conversation Envelope Diagrams}
The OVON (Open Voice Network) Conversation Envelope is a universal JSON structure whose purpose is to allow human or automatic agents (assistants) to interoperably participate in a conversation.

\begin{figure}[h!] 
	\centering
	\includegraphics[scale=0.06]{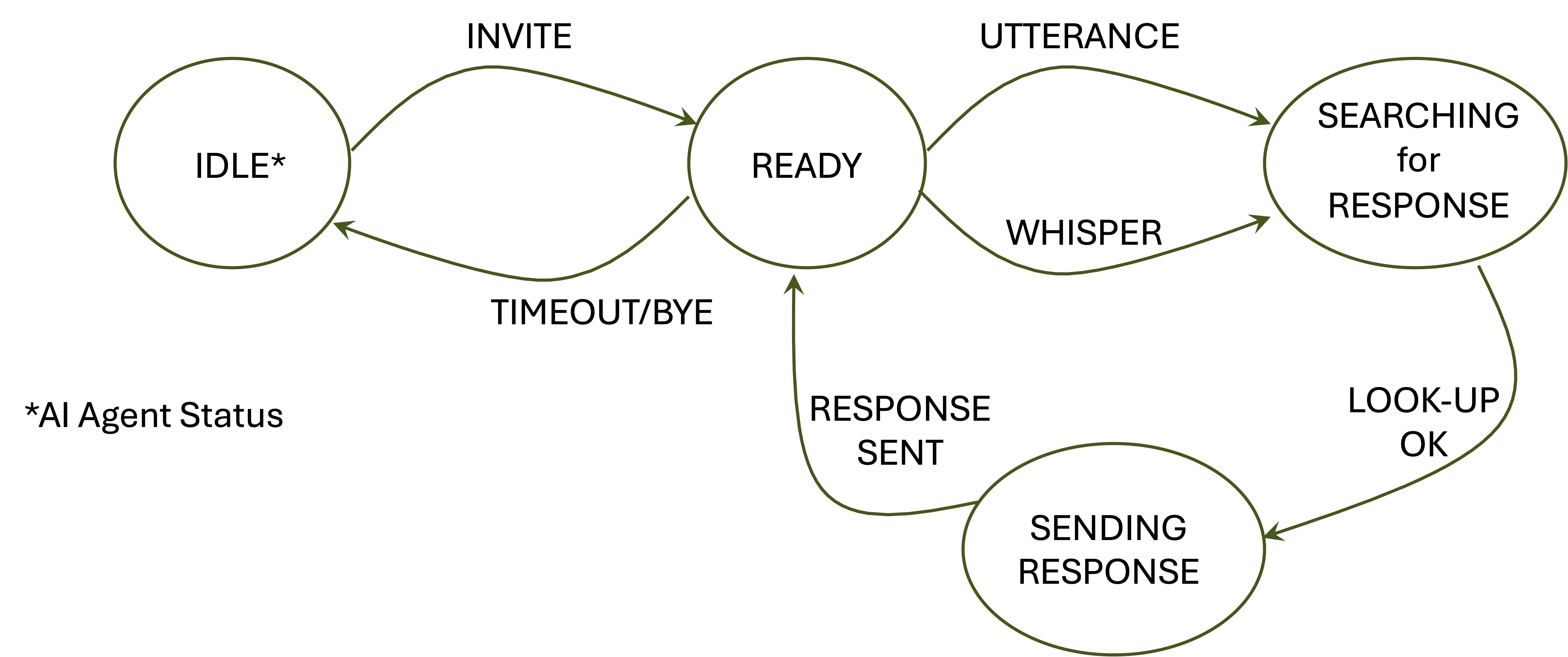}
	\caption{States and events related to a serving agent when it is successful in finding a response}
	\label{fig:fig1}
\end{figure}

The state diagram above (see Figure\ref{fig:fig1}) depicts the different states and transitions related to a serving agent in a normal conversation where the agent has been able to provide a response to an utterance or a whisper Conversation Envelope.\\
As shown in Listing~\ref{lst:post_request_annex_a}, the JSON structure illustrates an example of a POST request sent to an AI agent specializing in information about books and authors. For simplicity, this agent is referred to as 'Smartlibrary,' and the example includes both Utterance and Whisper events.
\begin{figure}[h!] 
	\centering
	\includegraphics[scale=0.06]{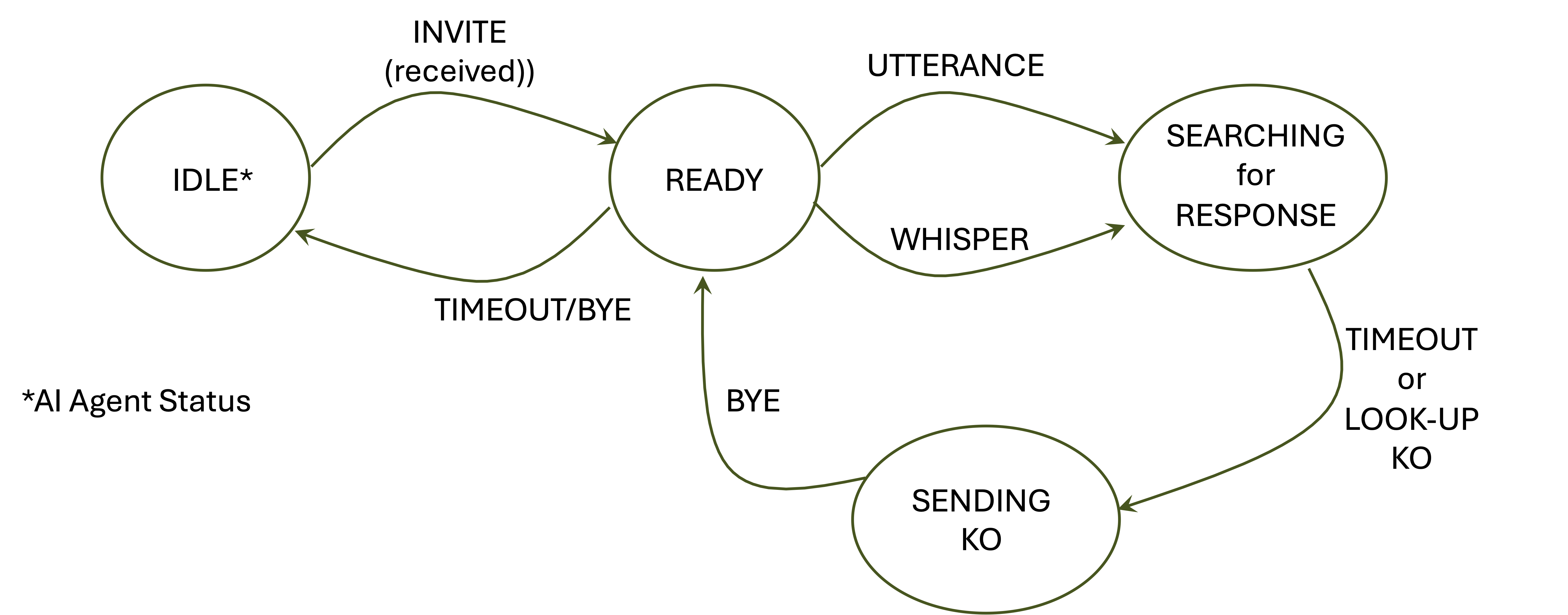}
	\caption{States and events related to a serving agent when it is unsuccessful in finding  a response}
	\label{fig:fig2}
\end{figure}
From the IDLE state, the serving agent moves to the READY state whenever it receives an INVITE standard message envelope event\cite{invite} from a demanding agent. While in the READY state, the agent can move to the Searching for Response state upon receiving an UTTERANCE or a WHISPER standard envelope event\cite{utterance} from the serving agent.
From the Searching for Response state, the agent can move to the Sending Response state whenever the query look-up is successful, moving back to the READY state upon sending the response to the demanding agent, so continuing to be ready for further utterance or whisper events. Finally, the serving agent can return to the IDLE state after proper inactivity timeout or a BYE event\cite{bye}.\\
The next state diagram (Figure\ref{fig:fig2}) represents the different states and transitions related to a serving agent when the agent has not been able to provide a response to an UTTERANCE or a WHISPER event. This can be due to various reasons, such as lookup failures, timeouts, or other situations.
Figure\ref{fig:fig1} and Figure\ref{fig:fig2} can be combined (see Figure\ref{fig:fig3}) to represent both use cases: the serving agent sending a response and the serving agent terminating the conversation without sending a response.

\begin{figure}[h!] 
	\centering
	\includegraphics[scale=0.06]{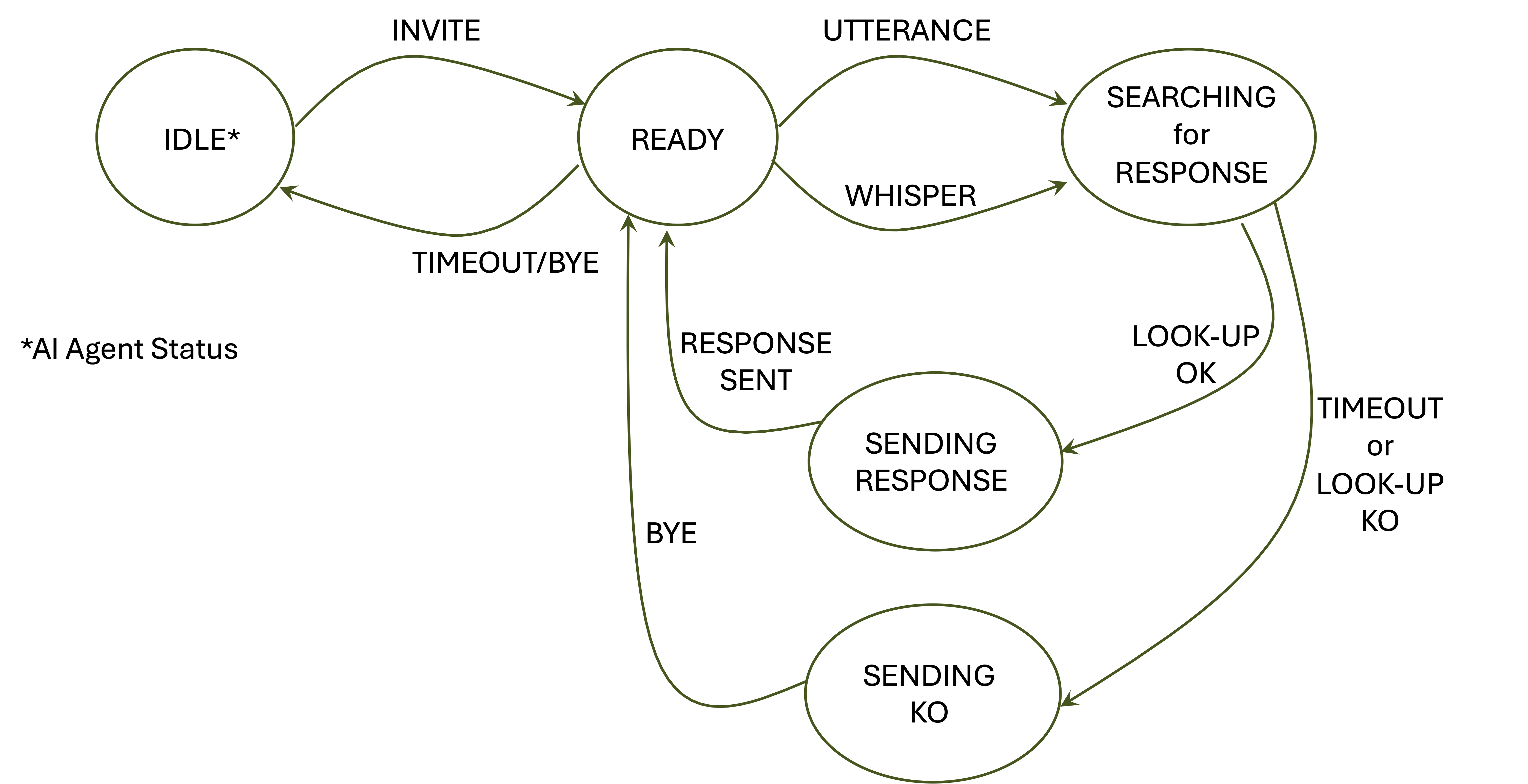}
	\caption{Combined states and events related to a serving agent}
	\label{fig:fig3}
\end{figure}

\begin{figure}[h!] 
	\centering
	\includegraphics[scale=0.06]{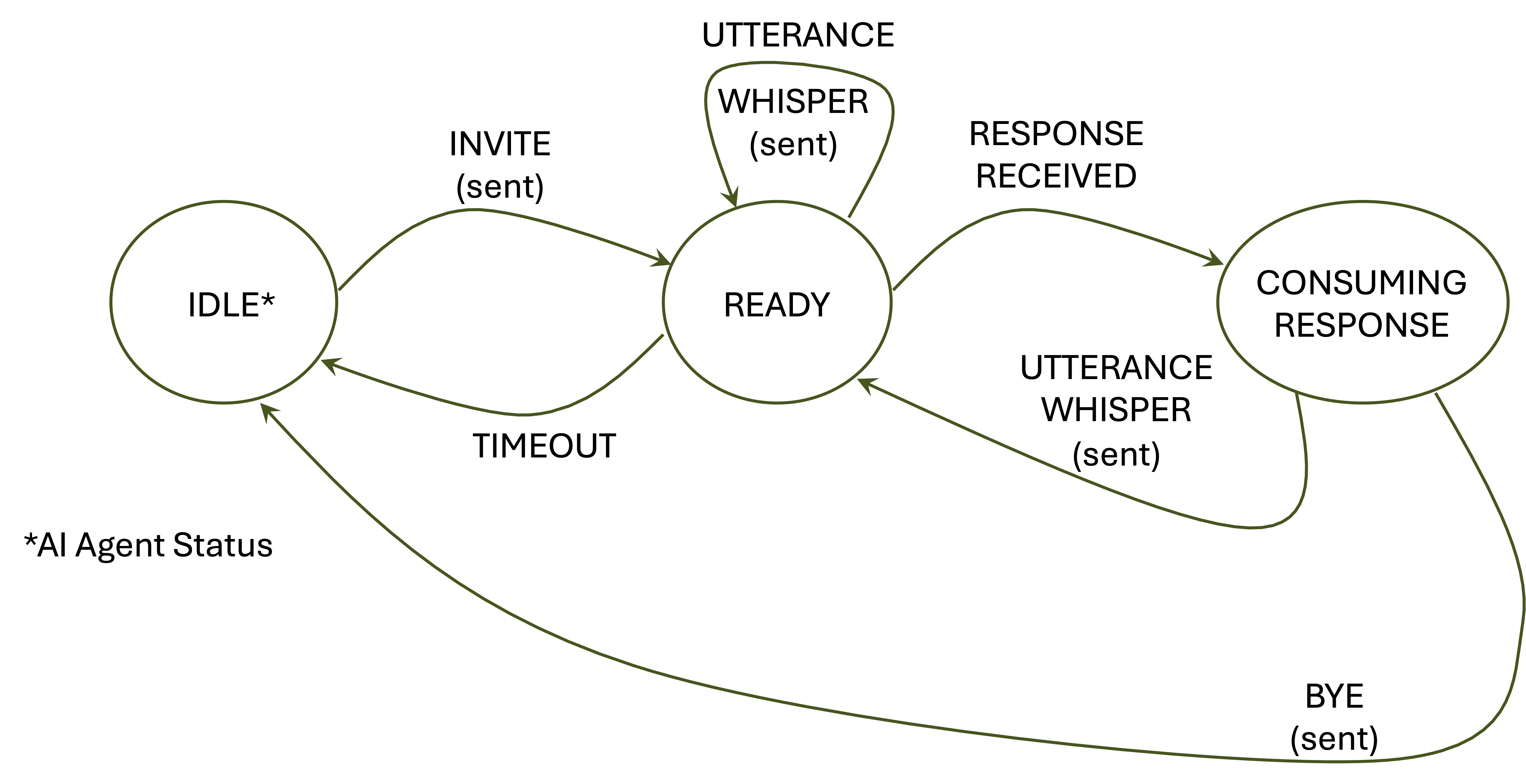}
	\caption{Different states and transitions related to a demanding agent}
	\label{fig:fig4}
\end{figure}

The state diagram in Figure\ref{fig:fig4} shows the different states and transitions related to a demanding agent. From IDLE state, the agent moves to the READY state whenever it sends an INVITE standard message envelope event\cite{invite}. While in the READY state, the agent can continue sending Utterances or Whispers according to the standard message envelope event\cite{utterance}, or it can move to the Consuming Response state upon receiving a response from the serving agent.

From the Consuming Response state, the demanding agent can move back to the READY state whenever it sends an Utterance or a Whisper event\cite{utterance}, or it can return to the IDLE state if the conversation has been terminated (BYE standard envelope event\cite{bye}).

\clearpage

\subsection{Interoperable Assistant Manifest Diagrams}
By using the Assistant Manifest specifications each agent can describe its capabilities to other agents. This provides the basis for a DISCOVERY process by which an agent can find agents with specific capabilities.
\begin{figure}[h!] 
	\centering
	\includegraphics[scale=0.06]{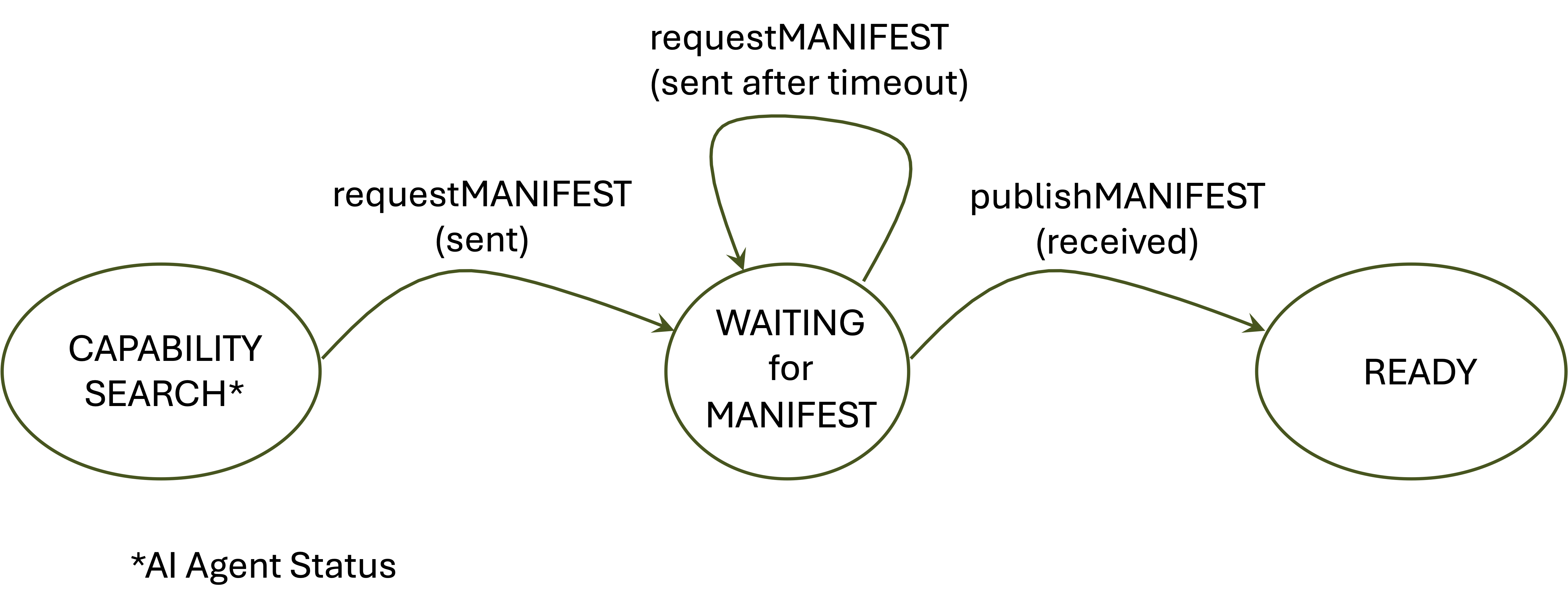}
	\caption{Demanding agent looking for the Manifest details of a target agent}
	\label{fig:fig5}
\end{figure}
The state diagram in Figure\ref{fig:fig5} shows the different states and transitions related to a demanding agent looking for the Manifest details of a target agent. From CAPABILITY SEARCH state, the agent moves to the WAITING for MANIFEST state whenever it sends a requestMANIFEST standard message envelope event\cite{reqmanifest}. While in the WAITING for MANIFEST state, the agent can continue sending requestMANIFEST periodically, or it can move to the READY state upon receiving a response from the serving agent via the publishMANIFEST standard message envelope event\cite{pubmanifest}.
The previous process is useful when the potential serving agent has already been identified by the demanding agent. In other cases, it is possible to search for a list of potential serving agents capable of accomplishing specific tasks using a natural language-based API similar to the Whisper events.
\begin{figure}[h!] 
	\centering
	\includegraphics[scale=0.06]{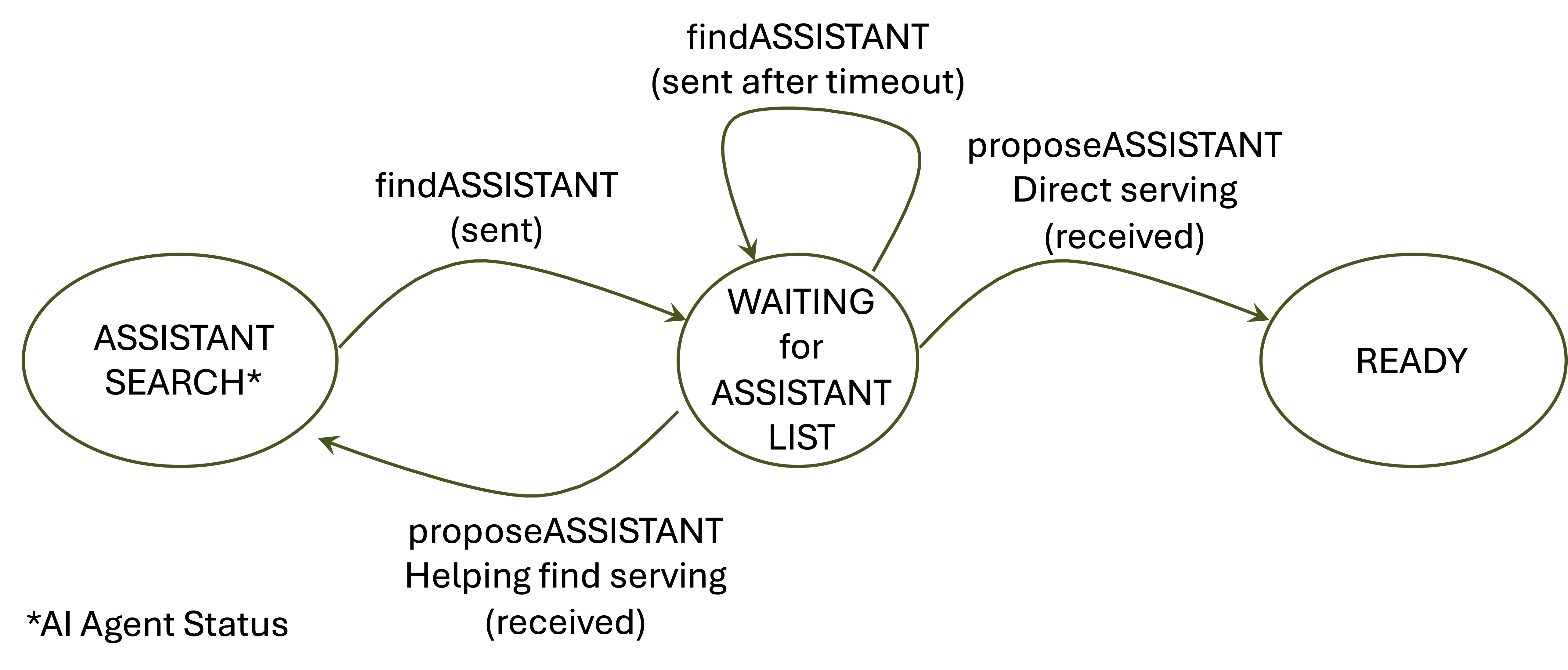}
	\caption{Demanding agent looking for Assistants available to perform specific tasks}
	\label{fig:fig6}
\end{figure}
The state diagram in Figure\ref{fig:fig6} shows the different states and transitions related to a demanding agent looking for Assistants available to perform specific tasks. From the ASSISTANT SEARCH state, the agent moves to the WAITING for ASSISTANT LIST state whenever it sends a findASSISTANT standard message envelope event\cite{findassistant}. While in the WAITING for ASSISTANT LIST state, the agent can continue sending findASSISTANT periodically, or it can move to the READY state upon receiving a response from the serving agent via the proposeMANIFEST standard message envelope event\cite{propassistant}, in case of recommendation involves one or more agents to directly service the request (including itself). In case the recommendation is related to one or more agents to help find who can service this request (indirect service), the agent will move back to the ASSISTANT SEARCH state and it will use the findASSISTANT event to query some of the suggested agents.

\clearpage

\section{USE CASES}

This section describes two scenarios in which multiple agents interact to coordinate and optimize a series of errands. This interaction is designed according to the Universal Open API specifications for Agentic Natural Language Multimodal Communications as described in this paper. 

\section*{Smart Errands Use Case}
In the first scenario, Emmett, a human, seeks assistance from Cassandra, his general AI assistant, to manage and streamline his possible errands efficiently. The AI assistants at various service points - Pat at Blooming Town Florist, Andrew at the Post Office, Charles at the hardware store, and Sukanya the Host at Thai Palace - facilitate the transactions. Emmett, a human, has the following goals:
\begin{itemize}
\item Order some flowers for his wife's birthday.
\item Check on the repair of the chainsaw he left at the hardware store.
\item Order some carryout Thai food for lunch.
\item Find the cost of mailing a 2 pound package to California.
\end{itemize}

\section*{Characters}

\begin{itemize}
	\item \textbf{Emmett}: The Human
	\item \textbf{Cassandra}: Emmett’s general AI assistant
	\item \textbf{Pat}: AI Assistant for his local florist
	\item \textbf{Andrew}: AI Assistant at the post office
	\item \textbf{Charles}: AI Assistant at Emmett's local hardware store
	\item \textbf{Sukanya}: AI Assistant at the local restaurant, “Thai Palace”
\end{itemize}

\section*{AI LLM Technologies}

\begin{itemize}
	\item \textbf{Cassandra}: based on GPT-3.5 Turbo
	\item \textbf{Pat}: llama2
	\item \textbf{Andrew}: rule-based application (no LLM involved)
	\item \textbf{Charles}: Claude.ai 3.5 Sonnet
	\item \textbf{Sukanya}: GPT-4o
\end{itemize}

\section*{Conversations}

\begin{verbatim}
Emmett: Hi Cassandra.
Cassandra: Hi Emmett! How can I assist you today?
Emmett: I do need to order some flowers for my wife's birthday.
Cassandra: Sure thing, Emmett! I'll connect you with the local florist shop.
[invites Blooming Town Florist to this conversation]
Pat: Hi Emmett! 
I'm Pat, your friendly florist. How can I help you with your flower selection today? 
Are you looking for something specific or need suggestions?
Emmett: Do you have any red Proteas?
Pat: Hi Emmett! Yes, we do have red Proteas. They're stunning and make a bold statement. 
Would you like to include them in your arrangement?
Emmett: Yes and add some eucalyptus in a clear vase, please.
Pat: Great choice, Emmett! 
Red Proteas with eucalyptus in a clear vase will look stunning. 
Would you like to add any other flowers or accents?
Emmett: No that's fine, send it to my home.
Pat: Great choice, Emmett! Shall I use the credit card on file for this order?
Emmett: Yes use the card on file.
Pat: Great choice, Emmett! Your red Proteas with eucalyptus in a clear vase will be sent 
to your home.  Thanks for your order! Have a blooming day!
[Pat returns a "bye" event and the conversation turns back to Cassandra]
Cassandra: Hi Emmett! How can I assist you today?
Emmett: I need to check with the hardware store to see if my chainsaw repair is finished.
Cassandra: Sure thing, Emmett! I'll check with the hardware store for you. 
One moment please.
[Cassandra invites Charles at the hardware store to join]
Charles: Hi Emmett! How can I help you today? 
Need anything from the store or checking on a repair?
Emmett: I want to know if my chainsaw repair is finished.
Charles: Hi Emmett! Let me check on that for you. Ah, yes, your chainsaw will be ready
by tomorrow afternoon. Anything else I can help you with?
Emmett: No that's all Thanks. Goodbye.
[Charles returns a "bye" event and the conversation turns back to Cassandra]
Cassandra: Hi Emmett! Your chainsaw repair will be ready by tomorrow afternoon. 
Thanks for checking in.
Cassandra: Hi Emmett! How can I assist you today?
Emmett: I want to order a carry out from my Thai restaurant.
Cassandra: Sure thing, Emmett! I'll connect you with your favorite Thai restaurant.
[Cassandra invites Thai Palace to join]
Sukanya: Hello, Emmett! I'm Sukanya, the Restaurant Manager. 
How can I help you with your carryout order today? 
Our special of the day is Pad Thai with shrimp. Would you like to try it?
Emmett: Yes I'll take one order of the pad Thai with shrimp and make it spicy.
Sukanya: Great choice, Emmett! Would you like to add our special appetizer
of the day, crispy spring rolls?
Emmett: Yes add 2 spring rolls.
Sukanya: Great choice, Emmett! Would you like to try our special of the day, 
Mango Sticky Rice for dessert? When would you like to pick up your order?
Emmett: No dessert today but I'd like to pick up my order in about an hour
Sukanya: Got it, Emmett! One spicy shrimp Pad Thai and two spring rolls. 
Ready for pick-up in an hour. 
Thank you and see you soon!
[Sukanya returns a "bye" event and the conversation turns back to Cassandra]
Cassandra: Hi Emmett! How can I assist you today?
Emmett: How much does it cost to mail a 2 LB package to California?
Cassandra: Hi Emmett! I can help with that. 
Let me connect you to the post office for accurate pricing.
[Cassandra invites the Post Office to join]
Andrew: Hi Emmett! I'm Andrew, the Postmaster. 
How can I assist you with sending mail and packages
through the United States Postal Service today?
Emmett: How much does a 2 LB package going to California cost?
Andrew: Hi Emmett! The cost depends on the shipping service you choose. 
For example, Priority Mail starts around $8.70. 
Rates can vary based on dimensions and delivery speed. Need more details?
Emmett: No that's good Thanks. Goodbye.
Andrew: Goodbye, Emmett.
[Andrew returns a "bye" event and the conversation turns back to Cassandra]
Cassandra: Hi Emmett! How can I assist you today?
Emmett: That's all I needed. Have a good day.
Cassandra: Thank you, Emmett! Have a wonderful day!
\end{verbatim}

\section*{OVON sequence diagram}

\begin{figure}[h!] 
	\centering
	\includegraphics[scale=0.08]{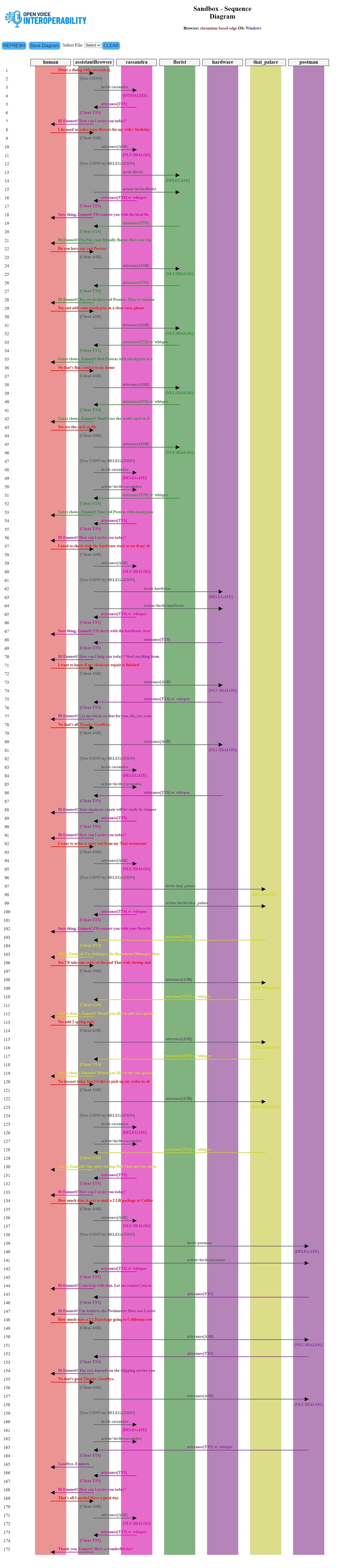}
	\caption{Interoperable AI Assistant scenario for Errands}
	\label{fig:fig7}
\end{figure}

As illustrated in APPENDIX C (Sequence Diagrams) on page \pageref{sec:sequence_diagrams}, Figure\ref{fig:fig7} shows a snapshot of the sequence related to the main OVON Conversation Envelope messages exchanged between the different Agents (Human and AI-based).

\section*{Summary}
In this use case, Emmett, a human, engages with Cassandra, his general AI assistant, and other AI assistants at various service points—his florist, the hardware store, the local Thai restaurant, and the Post Office. 
The scenario showcases how these AI agents efficiently manage tasks, demonstrating a seamless integration of technology that transcends specific AI models or frameworks. This is achieved independently of the underlying large language models (LLMs) and AI technologies deployed.

\section*{Smart Library Use Case}
The second use case illustrates an actual experiment conducted to provide potential conversational AI chatbot support to Estonian citizens\cite{burokratt}. It outlines a scenario where various AI assistants collaborate to provide support for a human user, Lea, who needs assistance with literature research. The AI assistants are characterized by distinct roles, different technologies and AI capabilities.

\section*{Characters}

\begin{itemize}
	\item \textbf{Lea}: the Human Citizen
	\item \textbf{Juri}: the Frontend AI assistant for Citizens
	\item \textbf{Andres}: AI Assistant with Discovery capabilities
	\item \textbf{Kaja}: a Smart Library AI assistant
\end{itemize}

\section*{AI LLM Technologies}

\begin{itemize}
\item \textbf{Juri}: GPT-3.5 Turbo
\item \textbf{Andres}: Mistral-7B-Instruct-v0.3
\item \textbf{Kaja}: HuggingFaceH4/zephyr-7b-beta
\end{itemize}

\section*{Conversations}

\begin{verbatim}
	Juri: Welcome to the OVON Smart Library service! 
	I can look up information about books if you provide a valid ISBN number
	with a Whisper. If you prefer, you can also send me a Natural Language utterance
	and I'll reply you!
	Lea: Hello this is Lea, I need to prepare a Literature essay. 
	Do you know about any books written by Lydia Koidula?
	Juri: Sure, Lea. Let me check with a proper specialized library assistant. 
	Hold on please.
	Juri to Andres: Can you send me a list of Assistants with knowledge about books
	related to Lydia Koidula please?
	Andres to Juri: Here it’s a list of Assistants:
	[proposedAssistant:
	Kaja, "url": https://smartlibrary1.ee
	Heli, "url": https://nationalLibraryArchive1.org
	Kalev, "url": https://estonianlibrary2.org]
	Juri to Kaja: Send me your Manifest please [requestManifest]
	Kaja to Juri: This is my Manifest 
	publishManifest:
	[{
		"identification":
		{
			"serviceEndpoint": "https://smartlibrary1.ee/smartlibrary",
			"organization": ”EE_example",
			"conversationalName": "smartlibrary",
			"serviceName": "Books and Authors AI agent",
			"role": "Book Specialist",
			"synopsis" : "AI agent specializing in providing information about 
			Estonian books and authors."
		},
		
		"capabilities": [
		{
			"keywords": [
			"books",
			"authors",
			"ISBN",
			"editors”…
			],
			"languages": [
			…
			]
		Juri to Kaja: Do you know about any books written by Lydia Koidula?
		Kaja to Juri: Lydia Koidula, the pen name for Lydia Emilie Florentine Jannsen, 
		was one of the most important figures in Estonian literature,
		particularly known for…
		Juri to Lea: Thank you for your patience. Lydia Koidula, the pen name for 
		Lydia Emilie Florentine Jannsen, was one of the most important figures in
		Estonian literature, particularly known for…
\end{verbatim}

\section*{OVON sequence diagram}

As illustrated in APPENDIX C (Sequence Diagrams), Figure\ref{fig:fig8} shows a snapshot of the sequence related to the main OVON Conversation Envelope messages exchanged between the different Agents (Human and AI-based).

\section*{Summary}
As can be seen in this second use case, when Lea asks for information on books by Lydia Koidula, Juri does not provide the information directly but instead coordinates with Andres to identify the appropriate assistant with better expertise. Andres suggests Kaja among others, and Juri interacts with Kaja to retrieve the necessary information, which is then relayed back to Lea. Also this scenario underscores that the AI framework's ability to provide such coordinated, specialized support is independent of the specific large language models (LLMs) and conversational AI technologies deployed. This independence suggests a versatile and adaptable AI system capable of integrating various technologies to meet specific user needs.

\section{IMPROVEMENTS AND FUTURE DIRECTIONS}
While the interoperability specifications are currently able to support many practical use cases, there are several future enhancements that we believe will extend their applicability to even more use cases. 
\begin{itemize}
\item The ability to represent multimodal interchanges among agents
\item Support for multi-party conversations
\item Formats to enable agents to exchange background information, conversation history, and other context in addition to users’ utterances
\item Redaction of sensitive data
\end{itemize}
Furthermore, to address the complexities of Conversational AI Security, Ethics, and Accountability within the proposed interoperability specifications, several additional enhancements can be considered for future directions. These improvements are based on the work of a related project, the Open Voice TrustMark initiative\cite{trustmark}, and will complement the specifications by strengthening the framework's capability to handle security risks, ensure ethical interactions, and maintain clear accountability.
Here are some potential future guidelines for enhancements:
\begin{itemize}
\item Enhanced Authentication, Authorization, and Accounting (AAA) Framework: 
Implement robust multi-factor authentication methods to verify the identity of all interacting agents.
Establish comprehensive authorization protocols to control access based on agent roles and responsibilities.
Introduce detailed accounting logs that capture all interactions for audit purposes, ensuring traceability and non-repudiation.
\item Bias Mitigation Mechanisms:
Develop specifications to detect and mitigate bias in AI responses, particularly focusing on language models' training data and output.
Include periodic reviews of AI systems for bias across different demographics to ensure fairness and inclusivity in AI-mediated interactions.
\item Transparency Guidelines:
Mandate the disclosure of AI decision-making processes, allowing users to understand how responses are generated.
Implement mechanisms that explain AI responses in understandable terms, particularly in high-stake scenarios.
\item Ethical Interaction Protocols:
Create ethical guidelines that AI agents must adhere to during interactions, including respect for user privacy, adherence to social norms, and prohibition of deceptive practices.
Enforce these guidelines through regular compliance checks and updates in response to emerging ethical considerations.
\item Accountability Tracking:
Develop a system to assign and track accountability for decisions made by AI agents, ensuring that all actions can be traced back to specific algorithms or data used.
Set up independent review boards that can assess AI actions and intervene in cases of ethical violations or disputes.
\item Sensitive Data Redaction and Encryption:
Integrate advanced data redaction tools to automatically identify and mask sensitive information within conversations.
Employ state-of-the-art encryption techniques to protect data integrity and confidentiality during transmission between agents.
\end{itemize}
By incorporating these enhancements into the interoperability specifications, we can significantly improve the security, ethics, and accountability of Conversational AI platforms, ensuring they are safer, fairer, and more reliable for all users.

\section{CONCLUSIONS}
The proposed approach based on a Universal set of Open APIs for Agentic Natural Language Multimodal Communications, utilising the Open Voice Conversation Envelope API, offers significant advancements in enabling independent conversational assistants to collaborate. The key benefits of this approach include its very loose coupling of assistants, which allows for a high degree of flexibility and ease of integration. It is agnostic to the underlying technologies on which agents are based, thereby supporting a diverse range of technologies, including Generative AI LLMs, LMM, LAM, Non-Generative AI, and even non-AI systems, across multiple communication channels such as voice, video, and text. The approach imposes minimal requirements on cooperating assistants, thus reducing the complexity of adding new ones. By relying on natural language interpretation, it simplifies the communication process between assistants. Additionally, it can seamlessly send information to human agents, enhancing its versatility in handling user requests. These features collectively represent a step forward in the development of interoperable and flexible conversational AI systems.\\
The state diagrams and use cases previously described illustrate how the specifications are meticulously designed to remain independent of any specific conversational AI technology, including but not limited to large language models (LLMs). This design ensures wide compatibility and interoperability across various AI platforms and technologies.\\
The proposed architecture for interoperable Conversational AI assistants is designed to be replicable and accessible via open repositories. Everyone can access the OVON specifications\cite{ovonspec} and the Sandbox\cite{sandbox}, an experimental browser/assistant system that implements InteropDialogEventSpecs and supports OVON Envelopes. This system features text and speech interfaces, includes a local server for hosting browser-based applications, and provides a list of assistant servers for experimentation. Additionally, it offers a basic Python as well as LLM based assistant server frameworks for developing new assistants.\\
Furthermore, incorporating the proposed advancements into the interoperability specifications will profoundly elevate the capabilities of Conversational AI platforms in managing security, ethical interactions, and accountability. These advancements, previously detailed, include comprehensive authentication protocols, bias mitigation, transparency measures, and ethical guidelines, alongside enhanced data protection techniques. Implementing these measures ensures AI platforms not only meet current needs but are also prepared for broader and more complex applications, making them indispensable tools in a technology-driven future.

\section*{ACKNOWLEDGMENTS}
We express our sincere appreciation to the Open Voice interoperability\cite{ovoninter} Team (Linux Foundation AI \& Data Foundation) for their invaluable contributions and support in developing the Interoperable Standards, particularly to Jon Stine, Jim Larson and David Attwater, together with all the Burokratt team\cite{burokratt}. Their expertise, suggestions, and resources have been pivotal in shaping a model that is both ethically grounded and practically effective in real-world applications.

\bibliographystyle{unsrt}  


\newpage

\section*{APPENDIX A (Conversation Envelope)}
Conversation Envelope Object Structure:
\begin{lstlisting}[language=json, caption=Conversation Envelope Object Structure, label=lst:object_annex_a]
	{
		"ovon": {
			
			"schema": {
				"version": "0.9.2"      
				"url": "https://github.com/open-voice-interoperability/docs/tree/main/schemas/conversation-envelope/0.9.2/conversation-envelope-schema.json"
			},
			
			"conversation": {
				"id": "31050879662407560061859425913208"
			},
			
			"sender": {
				"from": "https://example.com/message-from",
				"reply-to": "https://example.com/reply-message-to"
			},
			
			"events": [
			{
				"to" : "intended recipient A",
				"eventType": "event type A",
				"parameters": {
					"parameter 1" : { parameter 1 values }  
					"parameter n" : { parameter n values }  
				}
			},
			{
				"to" : "intended recipient B",
				"eventType": "event type B",
				"parameters": {
					"parameter 1" : { parameter 1 values }
					"parameter n" : { parameter n values }   
				}
			},
			]
		}
	}
\end{lstlisting}

Conversation Envelope JSON POST REQUEST example (utterance and whisper events sent to a “Smart Library” AI Assistant specialized in Author and Book information):
\begin{lstlisting}[language=json, caption=HTTPS POST REQUEST using the OVON Universal APIs with Utterance and Whisper, label=lst:post_request_annex_a]
	{
		"ovon": {
			"schema": {
				"version": "0.9.2",
				"url": "https://openvoicenetwork.org/schema/dialog-envelope.json"
			},
			"conversation": {
				"id": "conv_1699812834794"
			},
			"sender": {
				"from": "https://organization_url_from",
				"reply-to": "https://organization_url_to"
			},
			"responseCode": 200,
			"events": [
			{
				"eventType": "invite",
				"parameters": {
					"to": {
						"url": "https://your-smartlibrary-url-here"
					}
				}
			},
			{
				"eventType": "utterance",
				"parameters": {
					"dialogEvent": {
						"speakerId": "humanOrAssistantID",
						"span": { "startTime": "2023-11-14 02:06:07+00:00" },
						"features": {
							"text": {
								"mimeType": "text/plain",
								"tokens": [ { "value": "Can I have some info about Harry Potter please?" } ]
							}
						}
					}
				}
			},
			{
				"eventType": "whisper",
				"parameters": {
					"dialogEvent": {
						"speakerId": "humanOrAssistantID",
						"span": { "startTime": "2023-11-14 02:06:07+00:00" },
						"features": {
							"text": {
								"mimeType": "text/plain",
								"tokens": [ { "value": "In particular can I get some info about harry potter and the philosopher's stone " } ]
							}
						}
					}
				}
			}
			]
		}
	}
\end{lstlisting}

\newpage

Conversation Envelope JSON POST RESPONSE example:
\begin{lstlisting}[language=json, caption=Smartlibrary AI Agent POST RESPONSE example to Utterance and Whisper, label=lst:post_response_annex_a]
{"ovon":{"schema":{"version":"0.9.2","url":"https://openvoicenetwork.org/schema/dialog-envelope.json"},"conversation":{"id":"conv_1699812834794"},"sender":{"from":"Smart
			Library
			APIs"},"responseCode":{"code":200,"description":"OK"},"events":[{"eventType":"utterance","parameters":{"dialogEvent":{"speakerId":"assistant","span":{"startTime":"2024-07-13T19:18:25.855Z"},"features":{"text":{"mimeType":"text/plain","tokens":[{"value":"\"Certainly! \\\"Harry Potter and the Philosopher's Stone\\\" is the first novel in the \\\"Harry Potter\\\" series written by British author J.K. Rowling. The title of the book in the United States was changed to \\\"Harry Potter and the Sorcerer's Stone.\\\" The book was first published by Bloomsbury in the UK on June 26, 1997, and later in the US by
								Scholastic on September 1, 1998.\\n\\nThe novel introduces Harry Potter, an\""}]}}}}}]}}
\end{lstlisting}

\newpage

\section*{APPENDIX B (Manifest and Discovery)}
Assistant Manifest Object Structure example: 

\begin{lstlisting}[language=json, caption=Smartlibrary Manifest example, label=lst:manifest_annex_b]
{
	"identification":
	{
		"serviceEndpoint": "your-smartlibrary-url-here/smartlibrary",
		"organization": "Your_Organization",
		"conversationalName": "smartlibrary",
		"serviceName": "Books and Authors AI agent",
		"role": "Book Specialist",
		"synopsis" : "AI agent specializing in providing information about books and authors."
	},
	
	"capabilities": [
	{
		"keywords": [
		"books",
		"authors",
		"ISBN",
		"editors"
		],
		"languages": [
		"en-us"
		],
		"descriptiveTexts": [
		"Authors and Books information"
		],
		"modalities": [
		"text"
		],
		"contentType": "application/json"
	}
	]
}
\end{lstlisting}

\newpage

Conversation Envelope JSON POST REQUEST example (requestManifest event sent to a “Smart Library” AI Assistant specialized in Author and Book information):

\begin{lstlisting}[language=json, caption=requestManifest POST REQUEST example, label=lst:post_requestmanifest_annex_b]
{
	"ovon": {
		"schema": {
			"version": "0.9.2"
		},
		"conversation": {
			"id": "31050879662407560061859425913208"
		},
		"sender": {
			"from": "https://someBot.com"
		},
		"events": [
		{
			"to": "https://your-smartlibrary-url-here",
			"eventType": "requestManifest"
		}
		]
	}
}
\end{lstlisting}

\newpage

Example of JSON RESPONSE to requestManifest:
\begin{lstlisting}[language=json, caption=HTTPS POST REQUEST using the OVON Universal APIs with Utterance and Whisper, label=lst:post_manifestresponse_annex_b]
{
	"ovon": {
		"schema": {
			"version": "0.9.2"
		},
		"conversation": {
			"id": "31050879662407560061859425913208"
		},
		"sender": {
			"from": "https://your-smartlibrary-url-here",
			"to": "https://someBot.com"
		},
		"events": [
		{
			"eventType": "publishManifest",
			"parameters": {
				"manifest": {
					"identification": {
						"serviceEndpoint": "https://your-smartlibrary-url-here/smartlibrary",
						"organization": "Your_Organization",
						"conversationalName": "smartlibrary",
						"serviceName": "Books and Authors AI agent",
						"role": "Book Specialist",
						"synopsis": "AI agent specializing in providing information about books and authors."
					},
					"capabilities": [
					{
						"keywords": [
						"books",
						"authors",
						"ISBN",
						"editors"
						],
						"languages": [
						"en-us"
						],
						"descriptiveTexts": [
						"Authors and Books information"
						],
						"modalities": [
						"text"
						],
						"contentType": "application/json"
					}
					]
				}
			}
		}
		]
	}
}
\end{lstlisting}

\section*{APPENDIX C (Sequence Diagrams)}
\label{sec:sequence_diagrams}

\clearpage

\begin{figure}[h!] 
	\centering
	\includegraphics[scale=0.09]{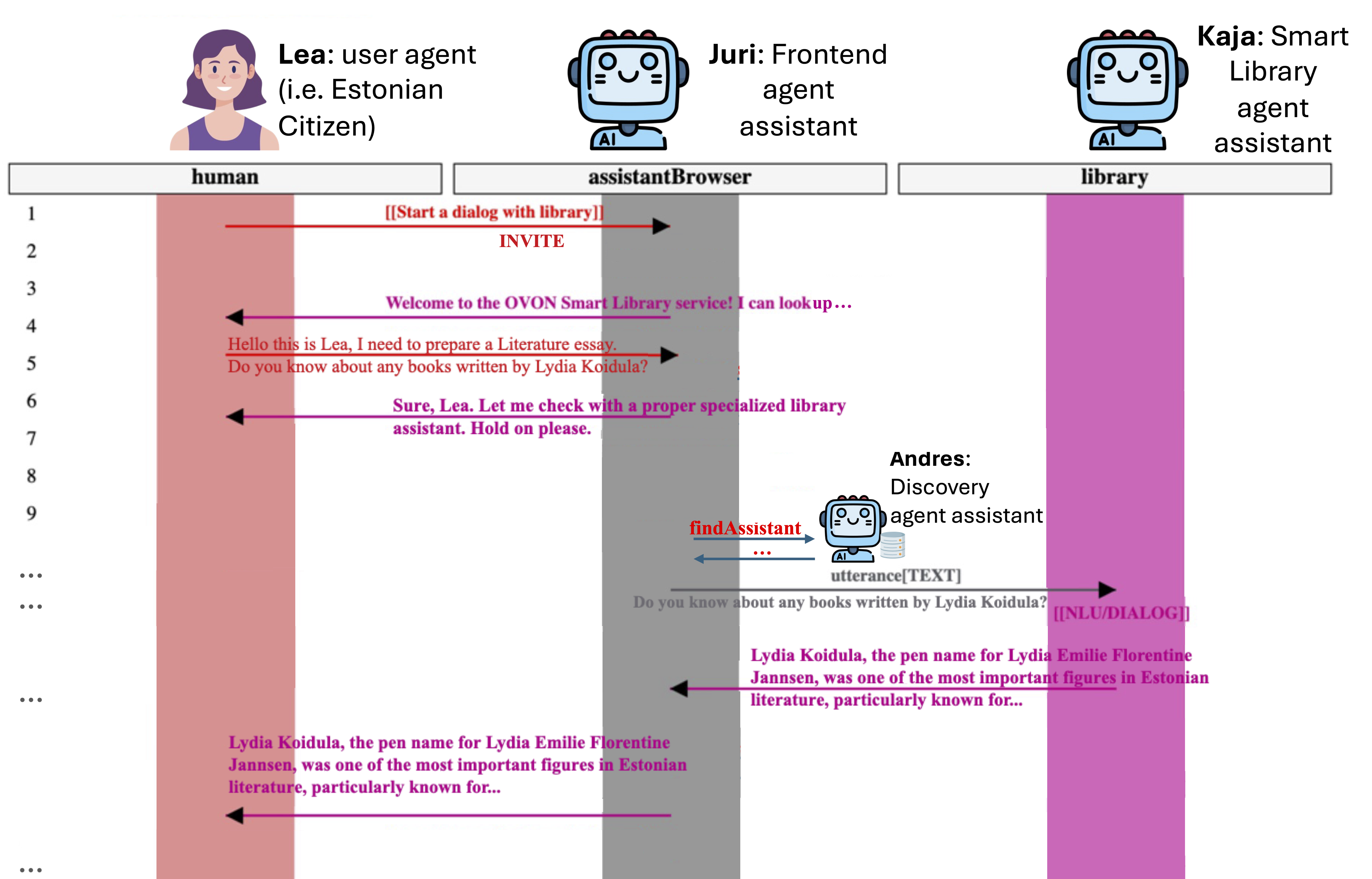}
	\caption{Interoperable Smart Library AI Assistant scenario for Citizens}
	\label{fig:fig8}
\end{figure}

Sequence Diagrams can be generated by running the Sandbox environment available in this repository\cite{sandbox}.

\end{document}